\documentclass[11pt]{article}

\PassOptionsToPackage{hyperfootnotes=false}{hyperref}
\usepackage[preprint]{acl}

\usepackage{times}
\usepackage{latexsym}

\usepackage[T1]{fontenc}

\usepackage[utf8]{inputenc}

\usepackage{microtype}

\usepackage{inconsolata}

\usepackage{graphicx}
\usepackage{pgfplots}
\pgfplotsset{compat=1.18}

\title{EvoGUI: An Evolution-Aware Benchmark for GUI State-Transition Understanding}
\hypersetup{
  pdftitle={EvoGUI: An Evolution-Aware Benchmark for GUI State-Transition Understanding},
  pdfauthor={Yaohan Yang, Minglei Shi, Borui Zhang, Jie Zhou, Jiwen Lu}
}

\author{
  \textbf{Yaohan Yang\thanks{Equal contribution.}},
  \textbf{Minglei Shi\footnotemark[1]\thanks{Project leader.}},
  \textbf{Borui Zhang},
  \textbf{Jie Zhou},
  \textbf{Jiwen Lu\thanks{Corresponding author.}}\\
  Department of Automation, Tsinghua University\\
  \texttt{\{yangyaoh22,sml25\}@mails.tsinghua.edu.cn}
}

\usepackage{booktabs}
\usepackage{array}
\newcommand{\modelname}{EvoGUI}
\newcommand{\benchname}{EvoGUI-Bench}

\begin{document}
\maketitle
\begin{abstract}
GUI agents must reason about how actions transform interface states, but end-to-end success rates entangle this ability with perception, grounding, planning, and recovery. We introduce \modelname{}, a diagnostic framework that converts normalized GUI trajectories into three complementary visual question answering probes: temporal ordering, inverse action/value prediction, and contrastive one-step successor discrimination. Their labels are derived from trajectory order and logged actions, requiring no additional task-label annotation after trajectory normalization. We instantiate \benchname{} from Mind2Web and WebLINX, yielding 3,000 instances across 120 domains, and evaluate 28 vision-language model configurations zero-shot. The strongest model reaches only 60.4 EvoGain, while model scale and GUI specialization do not reliably predict performance. These results establish \benchname{} as a scalable diagnostic complement to end-to-end GUI-agent evaluation while exposing substantial headroom in state-transition understanding. The source code is publicly available at \url{https://github.com/Yyhhh6/EvoGUI}.
\end{abstract}

\section{Introduction}

Vision-language models (VLMs) are increasingly used as the perception and reasoning backbone of graphical user interface (GUI) agents. Unlike static visual recognition, GUI interaction is inherently evolution-aware: an agent must understand not only what is visible on the current screen, but also how actions such as clicking, typing, or selecting transform the interface state. This transition-level understanding is necessary for reliable execution because agents must anticipate and interpret the consequences of their actions.

Existing GUI benchmarks have made substantial progress in evaluating agents in realistic web \cite{webarena,visualwebarena,workarena}, mobile \cite{androidworld}, and operating-system environments \cite{osworld,windowsagentarena}. These execution-based benchmarks are essential for measuring deployed task performance, but their success rates conflate perception, OCR, grounding, planning, tool use, and recovery. A transition-level diagnostic should therefore complement end-to-end success by exposing which local state changes a model can recognize and explain.

Recent dynamics-oriented evaluations use world modeling as a useful lens for diagnostic evaluation. ENACT \cite{enact} separates forward and inverse world modeling in embodied egocentric interaction, while MobileWorldBench \cite{mobileworldbench} studies semantic world modeling for mobile agents. These efforts suggest that transition-level evaluation can expose failures hidden by aggregate task success. For GUI agents, the missing step is a scalable way to turn existing trajectories into complementary transition probes while stating their perceptual and counterfactual limits explicitly.

\begin{figure*}[t]
\centering
\includegraphics[width=\textwidth]{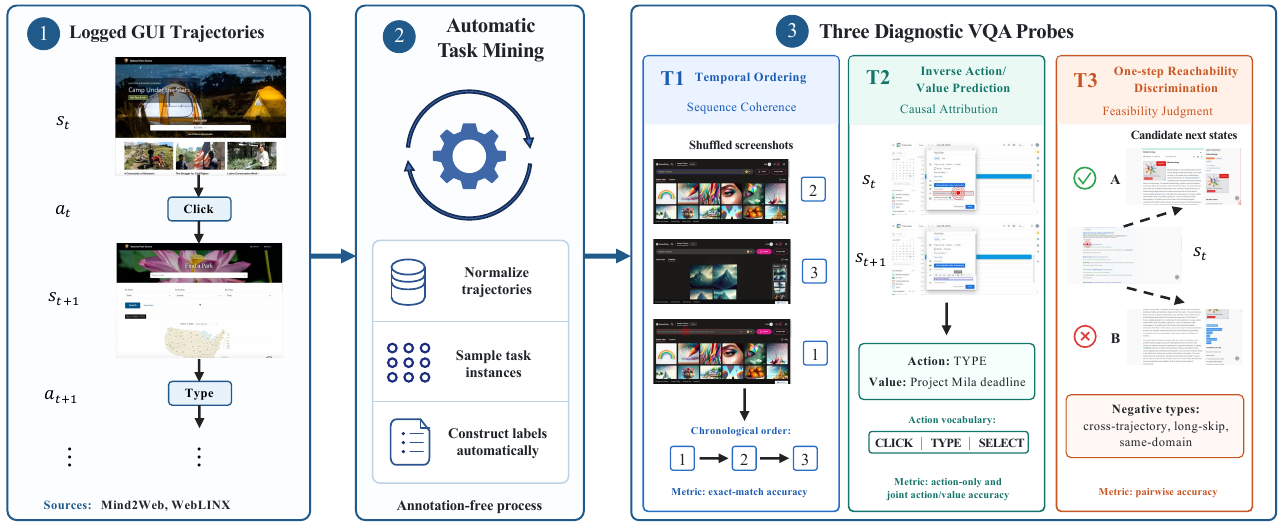}
\caption{Overview of \modelname{} benchmark construction. Logged GUI trajectories from Mind2Web and WebLINX are normalized into transition records and mined into three diagnostic VQA probes: T1 temporal ordering, T2 inverse action/value prediction, and T3 contrastive one-step successor discrimination. Labels are derived from trajectory order and recorded actions, requiring no additional task-label annotation after normalization; human inspection is used for quality control.}
\label{fig:evogui-overview}
\end{figure*}

In this work, we present \textbf{\modelname{}}, which turns normalized GUI trajectories into transition-level diagnostics without additional task-label annotation, and instantiate it as \textbf{\benchname{}}, a benchmark for GUI state-transition understanding. As illustrated in Figure~\ref{fig:evogui-overview}, \modelname{} factorizes this broad capability into temporal ordering, inverse action/value prediction, and contrastive one-step successor discrimination. T2 is reported in a relatively perception-light action-only form and a stricter joint form that also measures text/value recovery; T3 identifies the logged adjacent successor against a sampled distractor rather than proving that the distractor is impossible under every application state.

Because their labels come from logged trajectories, the construction can be transferred to additional GUI corpora through schema normalization and action-vocabulary expansion.

Our contributions can be summarized as follows:

\begin{itemize}
    \item \textbf{Trajectory-derived Framework.} We develop \textbf{\modelname{}}, a scalable framework that transforms normalized GUI trajectories into VQA-style diagnostics without additional task-label annotation.

    \item \textbf{Benchmark.} We introduce \textbf{\benchname{}}, a 3,000-instance evolution-aware diagnostic benchmark across 120 domains for evaluating GUI state-transition understanding.

    \item \textbf{Systematic and Validated Analysis.} We evaluate 28 representative VLM configurations and show that EvoGain remains far from saturation.

\end{itemize}

\section{Benchmark}

Figure~\ref{fig:evogui-overview} summarizes the \modelname{} construction pipeline. The remainder of this section formalizes the normalized trajectory schema, the three diagnostic VQA tasks, the trajectory-based mining procedure, quality control, and the resulting \benchname{} split.

\subsection{Trajectory Schema}

\modelname{} converts existing GUI interaction logs into diagnostic visual question answering tasks. We assume each source dataset can be normalized into a trajectory
\[
  \tau = [(s_t, a_t, v_t, s_{t+1}, m_t)]_{t=1}^{T},
\]
where $s_t$ and $s_{t+1}$ are consecutive screenshots, $a_t$ is the recorded user operation, $v_t$ is its optional value (e.g., typed text or a selected option), and $m_t$ stores metadata such as the trajectory id, domain, instruction, and source dataset. The source trajectories may themselves be human- or expert-generated; \modelname{} avoids new task-label annotation by deriving diagnostic targets mechanically after this normalization step.

\subsection{Diagnostic Tasks}

The benchmark contains three tasks that probe complementary aspects of GUI state-transition understanding. T1 asks whether observed states can be temporally ordered, T2 asks which logged action and value explain an adjacent state change, and T3 asks which candidate is the logged immediate successor.

T1, \textit{temporal ordering}, evaluates temporal state structure. Given $K \in \{3,4,5\}$ screenshots sampled from the same trajectory and shown in a shuffled order, the model must output their original chronological order.

T2, \textit{inverse action/value explanation}, evaluates whether a model can explain an observed transition. Given a pair $(s_t, s_{t+1})$, the model predicts an action class from $\{\textsc{click}, \textsc{type}, \textsc{select}\}$. We score both action-only accuracy and joint action/value accuracy, for which \textsc{type} and \textsc{select} additionally require the exact logged value.

T3, \textit{one-step reachability discrimination}, assesses whether a model can select the logged adjacent successor $s_{t+1}$ from two shuffled candidates. The distractor is sampled from one of three tiers: \textit{cross-trajectory}, a screen from a different trajectory; \textit{same-domain}, a screen from the same domain or application; and \textit{long-skip}, a later screen $s_{t+n}$ from the same trajectory with $n \geq 3$. Long-skip distractors are especially diagnostic because they are real future states in the same workflow but not the recorded immediate successor. Because distractors are sampled from logs rather than execution-validated counterfactuals, T3 measures discrimination against the sampled candidate, not universal impossibility of reaching that screen in one action.

\subsection{Trajectory-Based Task Mining}

\modelname{} derives all three task labels from normalized trajectories. The miner uses only temporal order, recorded actions, action values already present in the source logs, and constructed candidate pairs; no new annotator judgment determines the benchmark labels.

For T1, the miner first selects trajectories with at least $K$ valid screenshots, samples temporally ordered index sets for $K$, shuffles the selected screenshots, and stores the chronological index order as the label.

For T2, the miner iterates over adjacent transitions and keeps steps whose normalized action belongs to the target vocabulary. Actions that carry semantic content are retained only when their value is available; this prevents a model from receiving credit for predicting \textsc{type} or \textsc{select} while ignoring the input value.

For T3, the miner treats a valid adjacent transition as the logged-successor pair and constructs a pairwise question by adding one sampled distractor.

\subsection{Human Quality Control}

Trajectory-derived labels can still yield visually ambiguous items because screenshots may include page loads, advertisements, asynchronous refreshes, or incidental updates. We therefore manually inspect candidate T3 pairs before finalizing the split and remove transitions whose visible difference is dominated by changes that cannot reasonably be attributed to the logged action. This human-in-the-loop step audits label quality without assigning new labels: retained targets remain mechanically determined by trajectory adjacency. 

\subsection{Benchmark Instantiation}

We instantiate the benchmark as \benchname{}, a 3,000-instance split constructed from Mind2Web \cite{mind2web} and WebLINX \cite{weblinx}. We map \textsc{click}/\textsc{hover} style operations to \textsc{click}, \textsc{type}/\textsc{textInput}/\textsc{paste} to \textsc{type}, and \textsc{select}/\textsc{change} to \textsc{select}. Other intents, such as \textsc{load}, \textsc{submit}, \textsc{scroll}, \textsc{copy}, and tab-management actions, are excluded from T2 but their screenshots can still contribute to T1 and T3.

The construction is not intrinsically browser-specific. After mobile or desktop logs are mapped to the same transition schema, their operations can be added to the action vocabulary---for example, \textsc{swipe}, \textsc{long-press}, and \textsc{scroll} for AndroidWorld \cite{androidworld}, or drag and desktop-specific operations for OSWorld \cite{osworld}. T1 and contrastive T3 retain the same construction logic, while T2 expands its normalized action set and value rules. Thus, extending \modelname{} requires schema normalization and action-vocabulary design rather than new task-label annotation, although each new source still requires domain-appropriate quality auditing.

\section{Experiments}

\subsection{Setup}

\paragraph{Models.}
We evaluate 28 complete vision-language model configurations. We group models by release status. The open-weight group includes Qwen3.5 \cite{qwen3.5}, Qwen3.6 \cite{qwen3.6-27b, qwen36_35b_a3b}, Qwen3-VL \cite{qwen3vl}, Gemma-3-4B-IT \cite{gemma3}, Kimi-K2.5/K2.6 \cite{kimi_k25, kimi_k26}, GLM-4.5V \cite{glm45v}, UI-TARS-1.5-7B \cite{qin2025ui}, and MiMo-V2.5 \cite{mimo25}. The closed-weight group includes Claude-3-Haiku \cite{anthropic_claude3haiku}, OpenAI GPT-5/GPT-5.4 variants \cite{openai_gpt5, openai_gpt54, openai_gpt54_mini, openai_gpt54_nano}, Doubao Seed-2.0 variants \cite{doubao_seed2}, GLM-5V-Turbo \cite{glm5v_turbo}, Gemini-3-Flash-Preview \cite{gemini3flash}, and Grok-4.20 \cite{grok420}.

All models are evaluated zero-shot with the same task prompts and JSON output parser.

\begin{table}[t]
\centering
\caption{Statistics of \benchname{}. Each task contains 1,000 instances.}
\label{tab:benchmark-stats}
\scriptsize
\setlength{\tabcolsep}{5pt}
\begin{tabular}{llr}
\toprule
Part & Slice & Count \\
\midrule
T1 & $K=3/4/5$ & 334 / 333 / 333 \\
T2 & click / type / select & 461 / 461 / 78 \\
T3 & cross / same-domain / long-skip & 334 / 333 / 333 \\
\midrule
All & unique domains & 120 \\
All & top-1 domain share & 3.5\% \\
All & median / p95 screens per traj. & 12 / 39 \\
All & WebLINX share & 49\% \\
\bottomrule
\end{tabular}
\end{table}

\paragraph{Data.}
We evaluate on \textsc{\benchname{}}, the 3,000-instance benchmark split summarized in Table~\ref{tab:benchmark-stats}. Each task contains 1,000 examples, and T2 uses the unified \{\textsc{click}, \textsc{type}, \textsc{select}\} action vocabulary. The split is balanced over T1 sequence length and T3 distractor type, and it is domain-balanced by construction.

\paragraph{Protocol and metrics.}
All model runs use temperature 0 and no cache reuse. For locally executed Qwen models, we use BF16 SGLang inference with one running request; hosted models use the corresponding provider endpoint with the same prompts and parser. Table~\ref{tab:main} standardizes the main comparison at \texttt{max\_image\_side=512}. We report T1 exact ordering accuracy, T2 action-only top-1 and action/value joint top-1, and T3 pairwise successor accuracy. T3 long-skip accuracy is also a meaningful logged-successor slice, which is reported in Table~\ref{tab:appendix-t3-slices} in Appendix. The random-choice baseline is the average of four random runs.

To synthesize these structurally decoupled probes with random baselines, we introduce the \emph{EvoGain Index}, a relative improvement score over the random baseline. For each task $i \in \{\mathrm{T1},\mathrm{T2},\mathrm{T3}\}$, we compute the normalized gain
\[
\mathrm{NormScore}_i =
\frac{\mathrm{ModelScore}_i - \mathrm{Baseline}_i}{100 - \mathrm{Baseline}_i} \times 100,
\]
where the model scores are T1 exact ordering accuracy, T2 joint accuracy, and T3 pairwise accuracy, respectively.
EvoGain is the macro-average of the three normalized gains. To quantify sampling uncertainty, we also perform instance-level bootstrap resampling with 1,000 resamples for each model and task and propagate the resampled scores through the same normalization; Appendix~\ref{sec:appendix-uncertainty} reports the resulting 95\% confidence intervals.

\subsection{Main Results}

Table~\ref{tab:main} shows that \benchname{} produces clear and interpretable model separation. After removing task-specific chance effects, EvoGain ranks Gemini-3-Flash-Preview highest at 60.4\%, followed by Seed-2.0-lite-260428 at 50.7\%. Their bootstrap intervals, 58.16--62.66 and 48.45--53.06, respectively, remain separated. In contrast, several models with high raw T3 scores drop substantially after normalization, and many adjacent models in the middle of the ranking have overlapping intervals. Even the best model exceeds random choice by only 50.3 points on T1, 50.4 points on T2 joint accuracy, and 33.7 points on T3, indicating substantial remaining headroom.

The results further show that state-evolution ability cannot be reliably inferred from general capability, GUI specialization, or parameter count. Although strong generalist models lead overall, the GUI-specialized UI-TARS-1.5-7B reaches only 25.4 EvoGain, and larger Qwen models do not consistently outperform smaller variants. For example, Qwen3.5-397B-A17B scores below Qwen3.5-122B-A10B, while Qwen3-VL-235B-A22B scores below Qwen3.5-35B-A3B. This non-monotonic trend, visualized in Appendix Figure~\ref{fig:qwen-param-evogain}, suggests that transition-level understanding should be evaluated directly rather than inferred from model scale or general performance.

\begin{table}[t]
\centering
\caption{Main results on \benchname{} at \texttt{max\_image\_side=512}.
Rows are grouped by release status and ordered within each group from lower to higher EvoGain. All values are percentages. Bold indicates the best result; underline indicates the second-best.}
\label{tab:main}
\scriptsize
\setlength{\tabcolsep}{3.0pt}
\resizebox{\columnwidth}{!}{
\begin{tabular}{@{}lrrrrr@{}}
\toprule
Model & T1 $\uparrow$ & T2 top-1 $\uparrow$ & T2 joint $\uparrow$ & T3 $\uparrow$ & EvoGain $\uparrow$ \\
\midrule
\multicolumn{6}{l}{\textit{Baseline}} \\
Random choice        & 8.0 & 33.3 & 15.0 & 49.9 & 0.0 \\
\midrule
\multicolumn{6}{l}{\textit{Open-weight models}} \\
Qwen3.5-0.8B        & 8.1  & 49.6 & 47.4 & 57.7 & 17.9 \\
Gemma-3-4B-IT       & 8.8  & 50.3 & 37.8 & 67.3 & 20.8 \\
UI-TARS-1.5-7B      & 16.2 & 52.2 & 49.3 & 63.4 & 25.4 \\
Qwen3.5-2B          & 12.3 & 52.1 & 47.1 & 71.3 & 28.4 \\
Qwen3.5-4B          & 27.6 & 55.8 & 49.0 & 69.6 & 33.5 \\
Qwen3.5-9B          & 28.9 & 59.9 & 51.4 & 70.3 & 35.4 \\
GLM-4.5V            & 30.9 & 61.9 & 52.3 & 71.2 & 37.1 \\
Qwen3-VL-235B-A22B  & 30.9 & 64.1 & 55.1 & 70.2 & 37.5 \\
Qwen3.6-27B         & 29.2 & 59.1 & 55.1 & 72.6 & 38.5 \\
Qwen3.5-397B-A17B   & 34.4 & 64.7 & \underline{56.6} & 70.3 & 39.5 \\
Qwen3.5-27B         & 29.7 & 62.3 & 54.8 & 74.1 & 39.6 \\
Qwen3.5-35B-A3B     & 31.1 & 60.9 & 54.8 & 74.7 & 40.5 \\
Kimi-K2.5           & 33.4 & 61.9 & 52.8 & 76.3 & 41.6 \\
Qwen3.6-35B-A3B     & 33.3 & 61.9 & \underline{56.6} & 74.7 & 42.0 \\
MiMo-V2.5           & 34.2 & 59.8 & 54.5 & 76.1 & 42.4 \\
Qwen3.5-122B-A10B   & 32.8 & 63.4 & 56.4 & 77.8 & 43.8 \\
Kimi-K2.6           & 34.4 & 61.2 & 52.6 & 79.4 & 43.9 \\
\midrule
\multicolumn{6}{l}{\textit{Closed-weight models}} \\
Claude-3-Haiku        & 8.8  & 51.7 & 36.2 & 62.7 & 17.1 \\
GPT-5-nano             & 13.3 & 47.3 & 41.0 & 69.4 & 25.1 \\
GPT-5.4-nano           & 25.1 & 55.8 & 49.0 & 73.5 & 35.2 \\
GPT-5-mini             & 30.3 & 59.3 & 52.5 & 69.2 & 35.6 \\
Grok-4.20            & 27.6 & 60.6 & 49.9 & 72.3 & 35.7 \\
Seed-2.0-mini-260428  & 37.2 & 62.3 & 45.9 & 72.3 & 37.6 \\
GLM-5V-Turbo         & 31.9 & 60.0 & 54.0 & 76.9 & 41.9 \\
GPT-5.4              & 42.1 & 60.4 & 50.9 & 75.4 & 43.4 \\
Seed-2.0-pro-260215   & 44.4 & \underline{66.0} & 55.7 & 80.9 & 49.8 \\
Seed-2.0-lite-260428  & \underline{49.4} & 62.3 & 51.7 & \underline{81.9} & \underline{50.7} \\
Gemini-3-Flash-Preview & \textbf{58.3} & \textbf{75.0} & \textbf{65.4} & \textbf{83.6} & \textbf{60.4} \\
\bottomrule
\end{tabular}
}
\end{table}

\subsubsection{T1: Temporal Ordering}

T1 provides the clearest separation of high-level state-evolution ability. However, even the best model still makes over 40\% exact-ordering errors, indicating that recovering multi-screen temporal structure remains challenging and cannot be assumed from general vision-language capability.

\subsubsection{T2: Inverse Action/Value Prediction}

T2 shows that action recognition is easier than full transition explanation. Across models, action-only accuracy consistently exceeds joint action/value accuracy, indicating that models can often detect salient visual changes but fail to recover value-bearing details such as typed strings or selected options. This gap should not be interpreted as pure causal-reasoning failure: joint exact match also depends on OCR, visual acuity, and value binding. Replacing T2 joint with action-only accuracy in EvoGain yields nearly the same comparison (Pearson $r=0.991$; Spearman $\rho=0.977$ across 28 models), showing that the overall ranking is not driven by exact-value recovery alone (Appendix~\ref{sec:appendix-alt-evogain}).

\subsubsection{T3: One-Step Reachability Discrimination}

T3 shows that aggregate GUI plausibility can overstate transition reasoning. While stronger models achieve high aggregate pairwise accuracy, Appendix Table~\ref{tab:appendix-t3-slices} shows that many still struggle with long-skip distractors, which require distinguishing logged immediate successors from later states in the same workflow. This gap suggests that current VLMs often capture coarse workflow plausibility without reliably identifying the recorded one-step transition. Appendix Table~\ref{tab:appendix-resolution} further shows that higher screenshot resolution improves inverse prediction but does not consistently repair temporal successor discrimination.

\subsection{Diagnostic Validity}

\paragraph{Dependence on visual evidence.}
We test Qwen3.5-27B in a matched control under normal input, black images with the textual prompt retained, and metadata-only input. Removing visual content reduces T1 exact accuracy from 28.1\% to 7.4\%/6.9\%, collapses T1 Kendall $\tau$ from 0.305 to approximately zero, and lowers T3 from 70.8\% to 49.1\%/49.6\%, near its 50\% random baseline. T2 action accuracy falls less sharply, from 57.5\% to 43.1\%/42.1\%, consistent with residual action-frequency and language priors. The near-identical black-image and metadata-only conditions indicate that task metadata alone provides little shortcut; Appendix~\ref{sec:appendix-no-vision} gives the full control table.

\paragraph{Relation to deployed agent success.}
\modelname{} is intended as a diagnostic complement to end-to-end performance. For five models with publicly reported OSWorld results \cite{osworld,mobileagent35,glm45v,glm5v_paper}, OSWorld success and EvoGain have positive rank association (Spearman $\rho=0.90$; Kendall $\tau=0.80$). For the three models reported under the same OSWorld-Verified harness, the rankings are identical. This comparison is encouraging but only indicative because it has five observations, mixes reporting protocols for two models, and crosses from web-derived offline transitions to desktop execution; Appendix~\ref{sec:appendix-osworld} provides sources and values.

\section{Conclusion}

In this work, we introduced \modelname{}, a trajectory-derived framework that factorizes GUI state-transition understanding into temporal ordering, inverse action/value explanation, and one-step reachability discrimination. The construction requires no additional task-label annotation after trajectory normalization and complements rather than replaces end-to-end agent evaluation. Across 28 VLM configurations, EvoGain remains far from saturation, and neither model scale nor GUI specialization reliably predicts transition-level performance. No-vision controls support material dependence on screenshots, while an OCR-reduced aggregation preserves the overall ranking. Together, these results motivate evaluating action-conditioned interface dynamics explicitly instead of inferring them from aggregate task success.

\section*{Limitations}

\modelname{} is a transition-level diagnostic, not a replacement for deployed GUI-agent success. Its OSWorld comparison is based on only five overlapping models and partially mixed reporting protocols, so the observed positive association should be read as indicative rather than predictive or causal. The benchmark also requires no additional manual annotation only after trajectory normalization: it still depends on the quality and coverage of existing human/expert trajectories. Our current 3,000-instance instantiation is browser-centered and dominated by web actions; although the transition schema can extend to mobile and desktop operations such as scrolling, swiping, long-pressing, dragging, and desktop-specific commands, such extensions require action-vocabulary expansion, empirical validation, and domain-specific quality control.

The three probes also differ in what they measure. T2 joint accuracy is a conservative action/value transition diagnostic rather than a pure causal-reasoning score, because exact value prediction can mix state-change understanding with OCR, value binding, and visual text recovery, especially for TYPE and SELECT actions. T3 measures contrastive one-step reachability by selecting the logged adjacent successor against sampled distractors; it cannot prove that a distractor is unreachable under every hidden application state or policy. Although human inspection can remove visibly ambiguous transitions and asynchronous page changes, larger audits and execution-validated counterfactuals are needed. Future work should therefore broaden trajectory sources, report uncertainty and slice-level reliability more systematically, and pair scalable automatic construction with stronger human and executable validation.

\section*{Ethical Considerations}

\modelname{} is constructed from publicly released Mind2Web and WebLINX trajectories and does not introduce new human-subject data collection or additional task-label annotation. Human inspection is used only to filter ambiguous mined items. The benchmark uses screenshots, recorded actions, and trajectory metadata to create diagnostic evaluation instances. 
We follow the licenses of the source datasets and remove or mask credentials, personal identifiers, and other sensitive text before release.
The benchmark is intended for measuring GUI state-transition understanding, not for deploying agents to perform real-world actions.

\bibliography{custom}

\appendix

\section{Reproducibility Details and Additional Results}

\subsection{Split Construction}

This appendix reports construction and evaluation details that are secondary to the main argument but necessary for auditing the benchmark. We construct the reported split from normalized Mind2Web and WebLINX trajectories, then sample 1,000 instances for each diagnostic task. The main paper gives the aggregate split statistics; here we specify the eligibility rules that determine which logged records can become labels.

For T1, screenshots are sampled from valid trajectories with $K \in \{3,4,5\}$ and are shuffled before being shown to the model. For T2, an instance is retained only when the normalized action is in the shared \{\textsc{click}, \textsc{type}, \textsc{select}\} vocabulary; value-bearing actions are retained only when the typed string or selected option is available in the source log. For T3, the target pair is a valid adjacent screenshot transition, including transitions whose source action is outside the T2 vocabulary, and the distractor is drawn from a cross-trajectory, long-skip, or same-domain pool. This separation keeps T2 value-sensitive while allowing richer logged interactions to contribute to temporal and successor-discrimination probes.

Automatically derived labels do not remove the need for quality control. Before finalizing the T3 split, we manually inspect candidate pairs and exclude items whose visible change is dominated by page loading, advertisements, asynchronous refreshes, or incidental content changes that cannot reasonably be attributed to the logged action. The inspectors do not assign task labels: the retained target is still fixed by trajectory adjacency. Thus, our claim is specifically that construction requires no additional \emph{task-label annotation} after trajectory normalization, not that the source demonstrations or quality audit are human-free.

\subsection{Relation Among Diagnostic Tasks}
\label{sec:appendix-task-relations}

T1, T2, and T3 are related but not redundant because each probe blocks a different shortcut. T1 can reward recognizing workflow progress, but it does not require identifying the user action that produced each transition. T2 fixes the before--after pair and asks for the causal action/value explanation, but it does not test whether the model can order a longer state sequence. T3 fixes the source state and asks for the logged immediate successor; its long-skip distractors are deliberately plausible future screens, which prevents eventual workflow plausibility from being treated as the recorded next transition. Reporting the probes separately therefore avoids interpreting success on one transition cue as evidence for all forms of GUI state-evolution understanding.

\subsection{Evaluation Protocol and Parsing}

All main-comparison model rows use a single zero-shot prompt per task, temperature 0, no cache reuse, and the same task files at \texttt{max\_image\_side=512}. Each model row contains 1,000 raw predictions per task, for 3,000 predictions per configuration. The main comparison therefore contains 84,000 raw model predictions across 28 configurations. The random-choice baseline in Table~\ref{tab:main} is the average of four random runs, and only the average row is used in the paper. Table~\ref{tab:appendix-resolution} additionally reports a complete 1024-pixel Qwen sweep for the eight models that also have complete 512-pixel runs.

The parser first extracts a JSON object and validates the task-specific schema. T1 is valid only when the \texttt{order} field is a complete permutation of the shown image indices. T2 is valid only when \texttt{action} belongs to the task vocabulary; the \texttt{value} field is exact-matched for \textsc{type} and \textsc{select}, and is ignored for \textsc{click}. T3 is valid only when \texttt{answer} selects one of the candidate labels. If JSON parsing fails, the fallback parser performs only conservative substring matching for the discrete label; it does not recover free-form values, so the T2 joint metric is not inflated by malformed responses. Outputs that remain invalid after fallback are counted as incorrect.

\subsection{Bootstrap Uncertainty}
\label{sec:appendix-uncertainty}

We estimate sampling uncertainty by resampling benchmark instances with replacement 1,000 times independently for each model and task. For each resample, we recompute T1 exact accuracy, T2 joint accuracy, and T3 pairwise accuracy and propagate them through the chance-normalized EvoGain formula. We then take the 2.5th and 97.5th percentiles as a 95\% confidence interval. Table~\ref{tab:appendix-top10-evogain-ci} reports the top-10 EvoGain rows with their confidence intervals. Gemini-3-Flash-Preview obtains 60.44 [58.16, 62.66], compared with 50.69 [48.45, 53.06] for Seed-2.0-lite-260428, so the leading pair is clearly separated. By contrast, the intervals of several models ranked 4--10 overlap.

\begin{table}[!h]
\centering
\caption{Top-10 models by EvoGain with bootstrap 95\% confidence intervals. T1, T2 joint, and T3 are the task accuracies used to compute EvoGain. All values except rank are percentages.}
\label{tab:appendix-top10-evogain-ci}
\scriptsize
\setlength{\tabcolsep}{2.2pt}
\resizebox{\columnwidth}{!}{
\begin{tabular}{@{}rllrrrr@{}}
\toprule
Rank & Model & EvoGain & 95\% CI & T1 & T2 joint & T3 \\
\midrule
1 & Gemini-3-Flash-Preview & \textbf{60.44} & [58.16, 62.66] & \textbf{58.26} & \textbf{65.46} & \textbf{83.64} \\
2 & Seed-2.0-lite-260428 & \underline{50.69} & [48.45, 53.06] & \underline{49.44} & 51.67 & \underline{81.91} \\
3 & Seed-2.0-pro-260215 & 49.81 & [47.52, 52.11] & 44.49 & \underline{55.77} & 80.86 \\
4 & Kimi-K2.6 & 43.96 & [41.59, 46.34] & 34.47 & 52.60 & 79.40 \\
5 & Qwen3.5-122B-A10B & 43.76 & [41.56, 46.07] & 32.76 & 56.43 & 77.77 \\
6 & GPT-5.4 & 43.45 & [40.92, 45.87] & 42.12 & 50.91 & 75.45 \\
7 & MiMo-V2.5 & 42.39 & [39.92, 44.84] & 34.25 & 54.51 & 76.03 \\
8 & Qwen3.6-35B-A3B & 41.96 & [39.68, 44.40] & 33.30 & 56.59 & 74.67 \\
9 & GLM-5V-Turbo & 41.92 & [39.42, 44.29] & 31.95 & 53.92 & 76.93 \\
10 & Kimi-K2.5 & 41.63 & [39.27, 43.96] & 33.44 & 52.85 & 76.31 \\
\bottomrule
\end{tabular}}
\end{table}

\begin{table}[t]
\centering
\caption{Action normalization used when constructing \benchname{}. Actions outside the shared vocabulary are excluded from T2 but may still provide screenshots for T1 and T3.}
\label{tab:appendix-action-normalization}
\scriptsize
\setlength{\tabcolsep}{3.8pt}
\begin{tabular}{lll}
\toprule
Source action & Unified action & T2 use \\
\midrule
\texttt{click}, \texttt{hover} & \textsc{click} & retained \\
\texttt{type}, \texttt{textInput}, \texttt{paste} & \textsc{type} & retained with value \\
\texttt{select}, \texttt{change} & \textsc{select} & retained with value \\
\texttt{load}, \texttt{submit}, \texttt{scroll} & not mapped & excluded \\
\texttt{copy}, tab-management actions & not mapped & excluded \\
\bottomrule
\end{tabular}
\end{table}

\subsection{T2 Per-Class Recall}
\begin{table}[t]
\centering
\caption{T2 aggregate action top-1 accuracy and per-class recall on \benchname{} for the model rows in Table~\ref{tab:main}. Rows use the same release-status grouping and model order as Table~\ref{tab:main}. All rows use \texttt{max\_image\_side=512}; values are percentages. Bold indicates the best result; underline indicates the second-best.}
\label{tab:appendix-t2-recall}
\scriptsize
\setlength{\tabcolsep}{2.2pt}
\resizebox{\columnwidth}{!}{
\begin{tabular}{@{}lrrrr@{}}
\toprule
Model & T2 top-1 $\uparrow$ & Click $\uparrow$ & Type $\uparrow$ & Select $\uparrow$ \\
\midrule
\multicolumn{5}{l}{\textit{Open-weight models}} \\
Qwen3.5-0.8B        & 49.6 & \textbf{96.5} & 10.8 & 1.3 \\
Gemma-3-4B-IT       & 50.3 & 56.4 & 48.6 & 24.4 \\
UI-TARS-1.5-7B      & 52.2 & \underline{96.3} & 16.7 & 1.3 \\
Qwen3.5-2B          & 52.1 & 85.5 & 25.6 & 11.5 \\
Qwen3.5-4B          & 55.8 & 81.8 & 36.9 & 14.1 \\
Qwen3.5-9B          & 59.9 & 80.0 & 45.8 & 24.4 \\
GLM-4.5V            & 61.9 & 80.3 & 49.7 & 25.6 \\
Qwen3-VL-235B-A22B  & 64.1 & 87.6 & 45.6 & 34.6 \\
Qwen3.6-27B         & 59.1 & 94.8 & 30.6 & 16.7 \\
Qwen3.5-397B-A17B   & 64.7 & 88.5 & 49.0 & 16.7 \\
Qwen3.5-27B         & 62.3 & 88.9 & 43.8 & 14.1 \\
Qwen3.5-35B-A3B     & 60.9 & 91.5 & 37.1 & 20.5 \\
Kimi-K2.5           & 61.9 & 80.5 & 48.8 & 29.5 \\
Qwen3.6-35B-A3B     & 61.9 & 95.2 & 35.4 & 21.8 \\
MiMo-V2.5           & 59.8 & 92.0 & 33.2 & 26.9 \\
Qwen3.5-122B-A10B   & 63.4 & 91.3 & 43.6 & 15.4 \\
Kimi-K2.6           & 61.2 & 80.9 & 46.0 & 34.6 \\
\midrule
\multicolumn{5}{l}{\textit{Closed-weight models}} \\
Claude-3-Haiku      & 51.7 & 57.9 & 46.9 & 43.6 \\
GPT-5-nano          & 47.3 & 76.8 & 24.3 & 9.0 \\
GPT-5.4-nano        & 55.8 & 83.3 & 34.3 & 20.5 \\
GPT-5-mini          & 59.3 & 84.8 & 40.3 & 20.5 \\
Grok-4.20           & 60.6 & 73.1 & 53.8 & 26.9 \\
Seed-2.0-mini-260428 & 62.3 & 55.3 & \textbf{73.8} & 35.9 \\
GLM-5V-Turbo        & 60.0 & 89.4 & 37.7 & 17.9 \\
GPT-5.4             & 60.4 & 81.3 & 40.8 & \underline{52.6} \\
Seed-2.0-pro-260215 & \underline{66.0} & 78.1 & 58.6 & 38.5 \\
Seed-2.0-lite-260428 & 62.3 & 67.7 & 59.4 & 47.4 \\
Gemini-3-Flash-Preview & \textbf{75.0} & 82.0 & \underline{69.6} & \textbf{65.4} \\
\bottomrule
\end{tabular}}
\end{table}
Table~\ref{tab:appendix-t2-recall} shows why T2 joint accuracy is a stricter inverse metric than aggregate action top-1 accuracy. Click recall can be high even when type and select recall remain low, which means that action-only accuracy can be driven by visually salient click-like transitions while underrepresenting value-bearing failures. The \textsc{select} slice contains only 78 instances, however, so its per-class values are diagnostic observations rather than stable model-level conclusions.

\subsection{OCR-Reduced EvoGain}
\label{sec:appendix-alt-evogain}

T2 joint accuracy intentionally requires exact typed or selected values and consequently mixes transition explanation with OCR, visual text recovery, and value binding. To test whether this perception-heavy component determines the overall comparison, we construct \mbox{EvoGain-Action} by replacing T2 joint accuracy with action-only accuracy and replacing the corresponding random baseline from 15.0\% with 33.3\%. Across the 28 configurations in Table~\ref{tab:main}, EvoGain-Action is highly consistent with the original index (Pearson $r=0.991$; Spearman $\rho=0.977$). Gemini-3-Flash-Preview remains first, and the leading and broad performance tiers change only marginally. The joint metric remains useful as a conservative test of complete action/value explanation, but the central cross-model conclusion does not depend on exact-value recovery.

\subsection{T3 Slice Results}
\begin{table}[t]
\centering
\caption{T3 accuracy by distractor type for the 512-pixel model rows in Table~\ref{tab:main}. Cross-trajectory candidates test coarse mismatch rejection, same-domain candidates provide related-interface distractors, and long-skip candidates test logged-successor discrimination against real later states from the same trajectory. All values are percentages. Bold indicates the best result; underline indicates the second-best.}
\label{tab:appendix-t3-slices}
\scriptsize
\setlength{\tabcolsep}{2.2pt}
\resizebox{\columnwidth}{!}{
\begin{tabular}{@{}lrrr@{}}
\toprule
Model & Cross-traj. $\uparrow$ & Same-domain $\uparrow$ & Long-skip $\uparrow$ \\
\midrule
\multicolumn{4}{l}{\textit{Open-weight models}} \\
Qwen3.5-0.8B        & 66.8 & 55.6 & 50.8 \\
Gemma-3-4B-IT       & 92.2 & 65.8 & 43.8 \\
UI-TARS-1.5-7B      & 80.8 & 57.7 & 51.7 \\
Qwen3.5-2B          & 95.8 & 68.2 & 49.9 \\
Qwen3.5-4B          & 94.9 & 66.7 & 47.2 \\
Qwen3.5-9B          & 97.6 & 67.9 & 45.4 \\
GLM-4.5V            & 96.4 & 71.5 & 45.6 \\
Qwen3-VL-235B-A22B  & 97.3 & 71.5 & 41.7 \\
Qwen3.6-27B         & 97.0 & 71.8 & 49.0 \\
Qwen3.5-397B-A17B   & 95.8 & 66.4 & 48.6 \\
Qwen3.5-27B         & 97.3 & 73.9 & 51.1 \\
Qwen3.5-35B-A3B     & 98.5 & 74.5 & 51.1 \\
Kimi-K2.5           & 97.3 & 77.5 & 54.1 \\
Qwen3.6-35B-A3B     & 98.5 & 75.7 & 49.9 \\
MiMo-V2.5           & 97.0 & 76.0 & 55.3 \\
Qwen3.5-122B-A10B   & 98.8 & 75.7 & 58.9 \\
Kimi-K2.6           & 99.4 & 81.7 & 57.1 \\
\midrule
\multicolumn{4}{l}{\textit{Closed-weight models}} \\
Claude-3-Haiku      & 82.9 & 59.5 & 45.6 \\
GPT-5-nano          & 88.0 & 65.2 & 55.0 \\
GPT-5.4-nano        & 94.6 & 73.9 & 52.0 \\
GPT-5-mini          & 97.9 & 70.9 & 38.7 \\
Grok-4.20           & 99.4 & 73.0 & 44.4 \\
Seed-2.0-mini-260428 & 98.2 & 76.3 & 42.3 \\
GLM-5V-Turbo        & 98.5 & 78.4 & 53.8 \\
GPT-5.4             & 97.0 & 70.9 & 58.3 \\
Seed-2.0-pro-260215 & \textbf{100.0} & 82.6 & 60.1 \\
Seed-2.0-lite-260428 & 98.5 & \textbf{84.1} & \underline{63.1} \\
Gemini-3-Flash-Preview & \underline{99.7} & \underline{83.2} & \textbf{67.9} \\
\bottomrule
\end{tabular}}
\end{table}

Table~\ref{tab:appendix-t3-slices} explains why the main paper reports aggregate T3 in Table~\ref{tab:main} and keeps long-skip accuracy as an appendix diagnostic. Cross-trajectory distractors mostly test coarse mismatch rejection, while long-skip distractors are real later states from the same workflow. The latter therefore better tests whether a model distinguishes eventual plausibility from the logged immediate successor, and it remains substantially harder for most models. Since these candidates are sampled rather than validated through exhaustive execution, the result is contrastive evidence and does not prove that every rejected screen is unreachable in one action.

\subsection{Resolution Comparison}
\label{sec:appendix-resolution}

\begin{table}[t]
\centering
\caption{Resolution comparison for the eight Qwen models with complete 512- and 1024-pixel sweeps. Each cell reports 512$\rightarrow$1024 accuracy; Long denotes T3 long-skip accuracy. Values are percentages. Bold and underline mark the best and second-best values within each side of a metric column, respectively.}
\label{tab:appendix-resolution}
\scriptsize
\setlength{\tabcolsep}{2.4pt}
\resizebox{\columnwidth}{!}{
\begin{tabular}{@{}lrrrrr@{}}
\toprule
Model & T1 & T2 top-1 & T2 joint & T3 & Long \\
\midrule
Qwen3.5-0.8B      & $8.1{\rightarrow}6.3$   & $49.6{\rightarrow}54.1$ & $47.4{\rightarrow}48.9$ & $57.7{\rightarrow}62.7$ & $\underline{50.8}{\rightarrow}50.8$ \\
Qwen3.5-2B        & $12.3{\rightarrow}12.9$ & $52.1{\rightarrow}54.9$ & $47.1{\rightarrow}49.3$ & $71.3{\rightarrow}73.2$ & $49.9{\rightarrow}\underline{51.1}$ \\
Qwen3.5-4B        & $27.6{\rightarrow}31.3$ & $55.8{\rightarrow}59.7$ & $49.0{\rightarrow}52.7$ & $69.6{\rightarrow}74.1$ & $47.2{\rightarrow}49.0$ \\
Qwen3.5-9B        & $28.9{\rightarrow}37.2$ & $59.9{\rightarrow}66.2$ & $51.4{\rightarrow}58.4$ & $70.3{\rightarrow}73.4$ & $45.4{\rightarrow}45.1$ \\
Qwen3.5-27B       & $29.7{\rightarrow}34.2$ & $\mathbf{62.3}{\rightarrow}\mathbf{69.1}$ & $54.8{\rightarrow}\underline{62.6}$ & $\underline{74.1}{\rightarrow}73.8$ & $\mathbf{51.1}{\rightarrow}46.3$ \\
Qwen3.5-35B-A3B   & $\underline{31.1}{\rightarrow}\mathbf{41.0}$ & $60.9{\rightarrow}68.2$ & $54.8{\rightarrow}62.5$ & $\mathbf{74.7}{\rightarrow}\underline{75.6}$ & $\mathbf{51.1}{\rightarrow}50.2$ \\
Qwen3.6-27B       & $29.2{\rightarrow}35.2$ & $59.1{\rightarrow}67.2$ & $\underline{55.1}{\rightarrow}\underline{62.6}$ & $72.6{\rightarrow}72.3$ & $49.0{\rightarrow}47.5$ \\
Qwen3.6-35B-A3B   & $\mathbf{33.3}{\rightarrow}\underline{40.4}$ & $\underline{61.9}{\rightarrow}\underline{69.0}$ & $\mathbf{56.6}{\rightarrow}\mathbf{63.2}$ & $\mathbf{74.7}{\rightarrow}\mathbf{77.8}$ & $49.9{\rightarrow}\mathbf{53.8}$ \\
\bottomrule
\end{tabular}}
\end{table}

Table~\ref{tab:appendix-resolution} indicates a consistent gain for inverse prediction: all eight Qwen rows improve on T2 top-1 and T2 joint accuracy when increasing the maximum image side from 512 to 1024, consistent with T2's reliance on fine-grained text and value recovery. Ordering improves for seven of the eight rows but less uniformly, whereas aggregate T3 and long-skip successor discrimination are non-monotonic. Resolution alone therefore does not solve the temporal transition probes. We retain 512 pixels for the fair main cross-model table and treat the 1024-pixel results as a controlled analysis.

\subsection{No-Vision Shortcut Controls}
\label{sec:appendix-no-vision}

We evaluate whether labels can be recovered without visual content using a matched Qwen3.5-27B control. The normal condition supplies the original images and prompt; the black-image condition replaces every screenshot with a black image while retaining the full textual prompt; and the metadata-only condition retains only source, domain, and task metadata. This control was executed as a separate matched run, so its normal row is used only for within-control comparisons rather than to replace the main-table result.

\begin{table}[t]
\centering
\caption{Qwen3.5-27B shortcut controls. Black-image and metadata-only inputs reduce T1 and T3 to approximately random performance. Values except Kendall $\tau$ are percentages.}
\label{tab:appendix-no-vision}
\scriptsize
\setlength{\tabcolsep}{3.2pt}
\resizebox{\columnwidth}{!}{
\begin{tabular}{@{}lrrrr@{}}
\toprule
Setting & T1 exact $\uparrow$ & T1 Kendall $\tau\uparrow$ & T2 action $\uparrow$ & T3 $\uparrow$ \\
\midrule
Normal        & 28.1 & 0.305  & 57.5 & 70.8 \\
Black image   & 7.4  & 0.004  & 43.1 & 49.1 \\
Metadata only & 6.9  & $-0.002$ & 42.1 & 49.6 \\
\bottomrule
\end{tabular}}
\end{table}

Removing vision reduces T1 exact accuracy by 73.7--75.4\%, T2 action accuracy by 25.0--26.8\%, and T3 accuracy by 29.9--30.6\% relative to the matched normal condition. T1 Kendall $\tau$ collapses to approximately zero and T3 approaches its 50\% random baseline. Residual T2 performance is consistent with language priors, including the higher frequency of \textsc{click}, rather than evidence that T2 can be solved from metadata alone.

\subsection{Qwen Parameter Trend}
\label{sec:appendix-qwen-params}

\begin{figure}[t]
\centering
\begin{tikzpicture}
\begin{axis}[
    width=\columnwidth,
    height=0.68\columnwidth,
    xmode=log,
    log basis x=10,
    xmin=0.6,
    xmax=650,
    ymin=15,
    ymax=46,
    xlabel={Total parameters (B, log scale)},
    ylabel={EvoGain},
    tick label style={font=\scriptsize},
    label style={font=\scriptsize},
    legend style={font=\scriptsize, draw=none, fill=none, at={(0.97,0.03)}, anchor=south east},
    grid=major,
    grid style={dashed, gray!30},
]
\addplot+[thick, mark=o] coordinates {
    (0.8,17.9) (2,28.4) (4,33.5) (9,35.4)
    (27,39.6) (35,40.5) (122,43.8) (397,39.5)
};
\addlegendentry{Qwen3.5}
\addplot+[thick, mark=square*] coordinates {
    (27,38.5) (35,42.0)
};
\addlegendentry{Qwen3.6}
\addplot+[only marks, mark=triangle*] coordinates {
    (235,37.5)
};
\addlegendentry{Qwen3-VL}

\end{axis}
\end{tikzpicture}
\caption{Qwen-family total parameter count versus EvoGain on \benchname{}. The trend is not monotonic: Qwen3.5-397B-A17B scores below Qwen3.5-122B-A10B, and Qwen3-VL-235B-A22B scores below several smaller Qwen variants.}
\label{fig:qwen-param-evogain}
\end{figure}
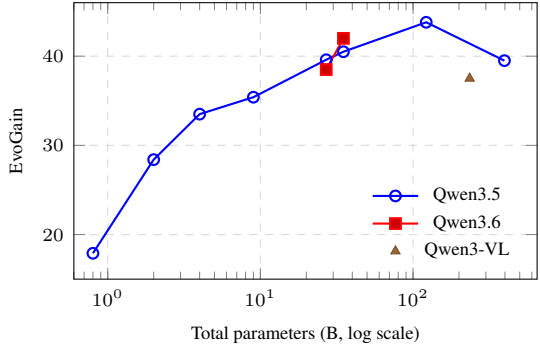

Figure~\ref{fig:qwen-param-evogain} supports the main-text claim that parameter count is an insufficient proxy for GUI state-evolution ability. Larger models often improve over smaller ones in the same family, but the relationship is not deterministic; architectural choices, training data, and transition-oriented supervision likely matter in addition to scale.

\subsection{Indicative Relation to OSWorld Success}
\label{sec:appendix-osworld}

We compare EvoGain with publicly reported OSWorld task-success rates for five overlapping models. We prioritize the shared OSWorld-Verified harness reported by Mobile-Agent-v3.5 \cite{mobileagent35} and supplement it with the official GLM-4.5V and GLM-5V-Turbo reports \cite{glm45v,glm5v_paper}. The underlying environment is OSWorld \cite{osworld}.

\begin{table}[t]
\centering
\caption{Indicative comparison with OSWorld success. The first, third, and fifth rows use the same OSWorld-Verified harness; the two GLM rows come from model-specific official reports. Values are percentages.}
\label{tab:appendix-osworld}
\scriptsize
\setlength{\tabcolsep}{4.2pt}
\begin{tabular}{@{}lrr@{}}
\toprule
Model & OSWorld $\uparrow$ & EvoGain $\uparrow$ \\
\midrule
UI-TARS-1.5-7B     & 27.4 & 25.4 \\
GLM-4.5V           & 35.8 & 37.1 \\
Qwen3-VL-235B-A22B & 38.1 & 37.5 \\
GLM-5V-Turbo       & 62.3 & 41.9 \\
Kimi-K2.5          & 63.3 & 41.6 \\
\bottomrule
\end{tabular}
\end{table}

Across all five models, the rankings have Spearman $\rho=0.90$ and Kendall $\tau=0.80$. Restricting the comparison to UI-TARS-1.5-7B, Qwen3-VL-235B-A22B, and Kimi-K2.5, whose OSWorld values use the same harness, yields identical rankings ($\rho=\tau=1.0$). This small cross-domain sample supports practical relevance only as an association: it cannot establish that EvoGain predicts agent success, and protocol heterogeneity may affect the five-model result.

\subsection{Interpretation Boundaries}

The appendix tables should be read as task-local diagnostics rather than calibrated measures of intrinsic task difficulty. T1, T2, and T3 use different answer formats, chance levels, and scoring rules, so absolute accuracies are not directly comparable across tasks. EvoGain normalizes chance performance for the main ranking, but it remains a diagnostic summary rather than a utility measure of deployed agent performance.

The additional slices refine the main claims without broadening them beyond the benchmark construction. T2 exact-value scoring tests whether the model can recover the logged typed string or selected option from the visual transition; errors may reflect failures in OCR, state comparison, or value binding. T3 distractors are sampled contrastive candidates from trajectory structure, so a correct answer demonstrates discrimination against the sampled candidate, not proof that the rejected screen is impossible under every application state or user policy. These boundaries support the narrower conclusions used in the main text: exact temporal ordering remains difficult, complete action/value explanation is limited by value recovery, and long-skip distractors make logged-successor discrimination harder than coarse cross-trajectory rejection.

\subsection{LLM Usage Statement}
Generative AI assistants were used for manuscript language polishing and partial code framework optimization. All outputs were manually reviewed and verified by the authors, who take full responsibility for the entire work.

\subsection{Prompt Templates}
\label{sec:appendix-prompts}

Tables~\ref{tab:appendix-prompt-t1}--\ref{tab:appendix-prompt-t3} report verbatim the complete zero-shot prompt templates used in the experiments. We typeset them as full-width role-separated tables without vertical rules, so that the model-facing instructions, instance-specific attachments, and parsed output fields are easy to audit. T1 parses \texttt{order}, T2 parses \texttt{action}/\texttt{value}/\texttt{confidence}, and T3 parses \texttt{answer}/\texttt{confidence}. Placeholders such as the user goal, candidate labels, and image slots are filled from each instance. The prompts ask the model to inspect the screens before answering, but the required output is a JSON object only; no free-form reasoning trace is parsed or evaluated. The T3 prompt uses ``reachable'' language, whereas our revised interpretation is restricted to identifying the logged adjacent successor against the sampled distractor.

\newenvironment{prompttable}[2]{
\begin{table*}[t]
\centering
\caption{#1}
\label{#2}
\footnotesize
\setlength{\tabcolsep}{5.5pt}
\renewcommand{\arraystretch}{1.12}
\begin{tabular}{@{}>{\raggedright\arraybackslash}p{0.16\textwidth}>{\raggedright\arraybackslash}p{0.78\textwidth}@{}}
\toprule
}{
\bottomrule
\end{tabular}
\end{table*}
}

\begin{prompttable}{Prompt template for T1 temporal ordering. The model receives shuffled screenshots from one session and returns their chronological order.}{tab:appendix-prompt-t1}
\textbf{Task} & T1: Temporal Ordering \\
\midrule
\textbf{System} & You are an expert evaluator of GUI screenshots. You will reason carefully about temporal causality between screens, and answer in the requested JSON format only. \\
\addlinespace[0.45em]
\textbf{User} & I will show you $K$ screenshots from the same user session, but presented in SHUFFLED order. Each screenshot is labeled by its position in the shuffled sequence. The task is to recover the chronological order, namely which screenshot happened first, second, and so on. Think step by step about UI affordances, what action would plausibly lead from one screen to the next, and the user's apparent goal when provided. Respond with a single JSON object of the form \texttt{\{"order": [\ldots]\}}, where positions are integers from 1 to $K$ and each position is used exactly once. Output only the JSON, no markdown, and no commentary. \\
\addlinespace[0.45em]
\textbf{Attachments} & Each shuffled screenshot is attached after its position label. \\
\addlinespace[0.45em]
\textbf{Parsed output} & \texttt{order}: a complete permutation of the shown screenshot indices. \\
\end{prompttable}

\begin{prompttable}{Prompt template for T2 inverse action/value prediction. The model receives two consecutive screenshots and predicts the action that caused the transition.}{tab:appendix-prompt-t2}
\textbf{Task} & T2: Inverse Action/Value Prediction \\
\midrule
\textbf{System} & You are an expert evaluator of GUI screenshots. Given two consecutive screens of a user session, infer the single user action that caused the change. Answer in the requested JSON format only. \\
\addlinespace[0.45em]
\textbf{User} & Below are two screenshots. The first is before the user's action, and the second is after the action. Pick the single user action that most likely caused the change from the provided options. If the action is \texttt{type}, predict the EXACT typed text. If the action is \texttt{select}, predict the EXACT option chosen from the dropdown or radio group. If the action is \texttt{click}, set the value to an empty string. Use the user goal when it is provided. Reason briefly about visible differences such as cursor, focus, new content, URL change, or dialog appearance before committing. Answer with \texttt{\{"action": "<one option>", "value": "<typed/selected text or empty>", "confidence": <0..1>\}}. Output only the JSON\@. \\
\addlinespace[0.45em]
\textbf{Attachments} & The before and after screenshots are attached in order. \\
\addlinespace[0.45em]
\textbf{Parsed output} & \texttt{action}: one option from the task vocabulary; \texttt{value}: the typed or selected value, or an empty string for \textsc{click}; \texttt{confidence}: a number in $[0,1]$. \\
\end{prompttable}

\begin{prompttable}{Verbatim prompt used for T3. Its model-facing wording asks for one-step reachability; the evaluation target is the logged adjacent successor.}{tab:appendix-prompt-t3}
\textbf{Task} & T3: One-Step Reachability Discrimination \\
\midrule
\textbf{System} & You are an expert evaluator of GUI screenshots. Given a starting screen and candidate next screens, decide which candidate could be reached from the starting screen by a single user action. Answer in the requested JSON format only. \\
\addlinespace[0.45em]
\textbf{User} & Below is a starting screen, followed by two candidate next screens labeled A and B. Exactly one of A and B is reachable from the starting screen by a SINGLE action, such as one click, one type, one scroll, or one back operation. The other is not reachable in one step. Use the user goal when it is provided. Think about which UI elements are visible and clickable in the starting screen, and what their immediate effect would be. Reject candidates that imply multiple navigation steps, a different page entirely, or a state that requires intermediate screens to appear first. Answer with \texttt{\{"answer": "A" | "B", "confidence": <0..1>\}}. Output only the JSON\@. \\
\addlinespace[0.45em]
\textbf{Attachments} & The starting screen and the two candidates are attached in order. \\
\addlinespace[0.45em]
\textbf{Parsed output} & \texttt{answer}: either \texttt{A} or \texttt{B}; \texttt{confidence}: a number in $[0,1]$. \\
\end{prompttable}

\end{document}